\documentclass{article}
\usepackage[font={small}]{caption}
\usepackage{amsmath}
\usepackage{cite}
\usepackage{arxiv}
\usepackage{graphicx}
\usepackage[utf8]{inputenc} 
\usepackage[T1]{fontenc}  
\usepackage{hyperref}    
\usepackage{url}      
\usepackage{booktabs}    
\usepackage{amsfonts}    
\usepackage{nicefrac}    
\usepackage{microtype}   
\usepackage{lipsum}
\usepackage{float}
\usepackage{subfig}

\usepackage{authblk}
\usepackage{xcolor}

\usepackage{setspace}
\title{Bibliography management: \texttt{natbib} package}
\date { } 
\title{Data Augmentation using Generative Adversarial Networks (GANs) for GAN-based Detection of Pneumonia and COVID-19 in Chest X-ray Images}

\author[1,2,*]{Saman Motamed}
\author[4]{Patrik Rogalla}
\author[1,2,3]{Farzad Khalvati}
\affil[1]{Institute of Medical Science, University of Toronto}
\affil[2]{Department of Diagnostic Imaging, Neurosciences and Mental Health, The Hospital for Sick Children}
\affil[3]{Department of Mechanical and Industrial Engineering, University of Toronto}
\affil[4]{University Health Network, Toronto, Ontario, Canada}
\affil[*]{sam.motamed@mail.utoronto.ca}

\newpage
\begin{document}
\maketitle
\newpage
\begin{abstract}
 Successful training of convolutional neural networks (CNNs) requires a substantial amount of data. With small datasets networks generalize poorly. Data Augmentation techniques improve the generalizability of neural networks by using existing training data more effectively. Standard data augmentation methods, however, produce limited plausible alternative data. Generative Adversarial Networks (GANs) have been utilized to generate new data and improve the performance of CNNs. Nevertheless, data augmentation techniques for training GANs are under-explored compared to CNNs. In this work, we propose a new GAN architecture for augmentation of chest X-rays for semi-supervised detection of pneumonia and COVID-19 using generative models. We show that the proposed GAN can be used to effectively augment data and improve classification accuracy of disease in chest X-rays for pneumonia and COVID-19. We compare our augmentation GAN model with Deep Convolutional GAN and traditional augmentation methods (rotate, zoom, etc) on two different X-ray datasets and show our GAN-based augmentation method surpasses other augmentation methods for training a GAN in detecting anomalies in X-ray images. 
\end{abstract}
\section{Introduction} 
In recent years, Convolutional Neural Networks (CNNs) have shown excellent results on several tasks using sufficient training data \cite{krizhevsky2012imagenet, he2015delving, he2016deep}. One of the main reasons for poor CNN performance and over-fitting on training data remains limited-sized datasets in many domains such as medical imaging. Improving the performance of CNNs can be achieved by using the existing data more effectively. Augmentation methods such as random rotations, flips, and adding various noise profiles have been proposed \cite{zhang2019image, hao2020comprehensive} as some methods of augmentation. Typical data augmentation techniques use a limited series of invariances that are easy to compute however (rotation, flips, etc), limited in the amount of new data they can generate.
\par Generative Adversarial Networks (GANs) \cite{goodfellow2016nips} have been used for data augmentation to improve the training of CNNs by generating new data without any pre-determined augmentation method. Cycle-GAN was used to generate synthetic non-contrast CT images by learning the transformation of contrast to non-contrast CT images \cite{sandfort2019data}. This improved the segmentation of abdominal organs in CT images using a U-Net model \cite{ronneberger2015u}. Using Deep Convolutional-GAN (DCGAN) \cite{radford2015unsupervised} and Conditional-GAN \cite{mirza2014conditional} to augment medical CT images of liver lesion and mammograms showed improved results in classification of lesions using CNNs ~\cite{frid2018gan, wu2018conditional}. Data Augmentation GAN (DAGAN) \cite{antoniou2017data} was able to improve the performance of basic CNN classifiers on EMNIST (images of handwritten digits), VGG-Face (images of human faces) and Omniglot (images of handwritten characters from 50 different alphabets) datasets by training DAGAN in a source domain and generating new data for the target domain. There has not been any study on data augmentation using GANs for training other GANs. The challenge with using a GAN to augment data for another GAN is that newly generated images with the trained generator of the GAN follow the same distribution as the training images, and hence there is no new information to be learned by another GAN that is trained on the original images combined with the newly generated (augmented) images.

\par In this paper, we propose Inception-Augmentation GAN (IAGAN) model inspired by DAGAN \cite{antoniou2017data} for the task of data augmentation that specifically improves the performance of another GAN architecture. We trained our proposed IAGAN on two chest X-rays datasets, one containing normal and pneumonia images and the other dataset containing normal, pneumonia and COVID-19 images. We showed that a trained IAGAN model can generate new X-ray images, independent of image labels, and improve the accuracy of generative models. We evaluated the performance of IAGAN model by training a DCGAN for anomaly detection (AnoGAN) \cite{schlegl2017unsupervised} and showed improved results in classifying pneumonia and COVID-19 positive cases with improved area under the receiver operating characteristic (ROC) curve (AUC), sensitivity, and specificity. We showed our trained IAGAN is able to generate new domain specific data regardless of the class of its input images. This allowed for an unsupervised data augmentation, in the case of not having labels for a subset of the images in the dataset. By training the same DCGAN model on the augmented data using traditional augmentation methods and generating new data using another DCGAN for the task of augmentation, we showed the ineffectiveness of these methods in successful augmentation of data for training a generative model compared to our IAGAN for detecting pneumonia and COVID-19 images.

\section{IAGAN Architecture}
\begin{figure}
 \centering
 \includegraphics[width=\textwidth]{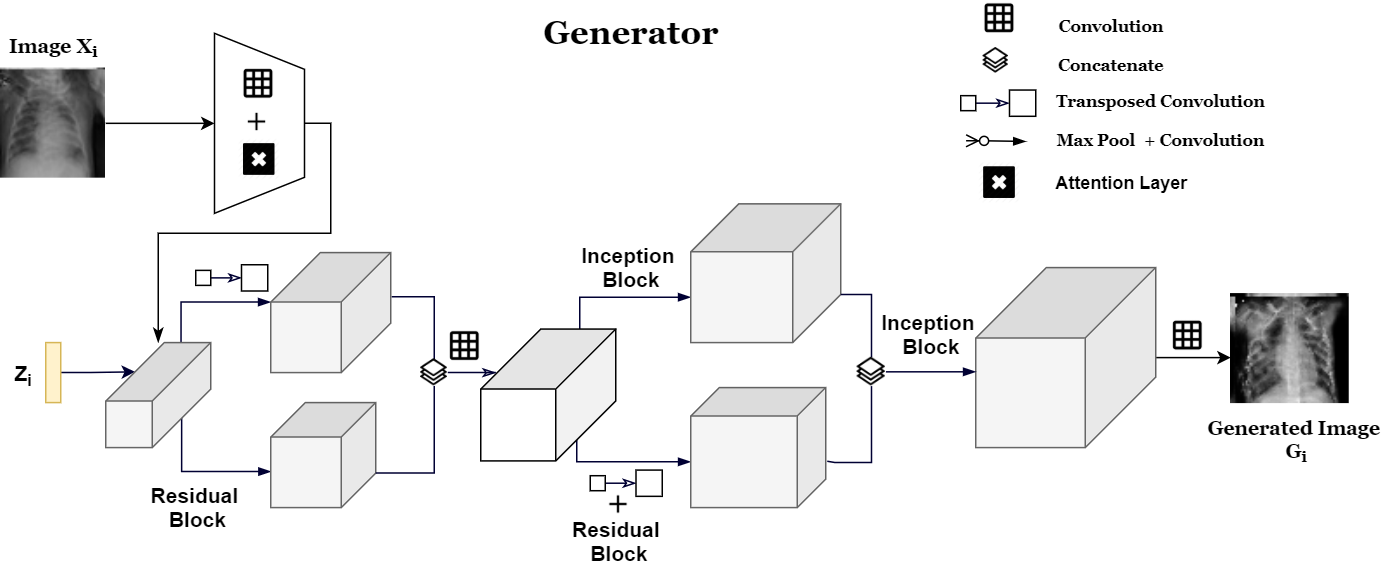}
 \caption{IAGAN's Generator Architecture}\label{fig1}
\end{figure}

\begin{figure}
 \centering
 \includegraphics[width=95mm]{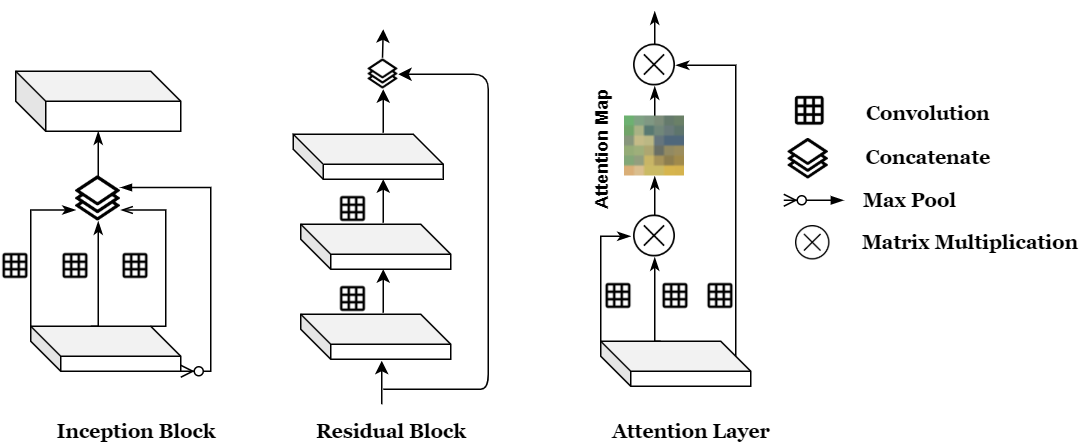}
 \caption{IAGAN's Generator specific architecture breakdown}\label{fig2}
\end{figure}

Figure~\ref{fig1} shows the architecture of the proposed IAGAN's Generator. At each iteration \(i\), as input, the generator (\textbf{G}) takes a Gaussian noise vector \(z_i\) and a batch of real training image \(x_i\). 
By encoding the input images \(x_i\) using convolution and attention layers to a lower-dimensional representation, before concatenating this representation of the image with the projected noise vector \(z_i\) (concatenation happens after \(z_i\) goes through a dense layer and non-linearity), we aim to not only use the full image representation using the discriminator, but also get a lower representation of images fed through the generator for better generalizability of \textbf{G} in generating images. The dual input to the generator also allows the trained generator to use images from different classes and generate a broader range of images to augment our specific training data class. The use of attention layers in GANs (Figure~\ref{fig2}) has shown to capture long-range dependencies in the image \cite{zhang2018self} where simple convolution layers focus on local features restricted by their receptive field, self-attention layers capture a broader range of features within the image. The attention layer uses three \(1 \times 1\) convolutions. \(1 \times 1\) convolution helps to reduce the number of channels in the network. Two of the convolution outputs, as suggested by Figure~\ref{fig2}, are multiplied (matrix multiplication) and fed to a \textit{softmax} activation, which results in producing the attention map. The attention map acts as the probability of each pixel affecting the output of the third convolution layer. Feeding a lower-dimensional representation of an input image \(x\) allows for the trained generator to use images from different classes to produce similar never-before-seen images of the class it was trained on.
\par Using inception and residual architectures \cite{szegedy2016rethinking} increase GAN's ability to capture more details from training image-space without losing spatial information after each convolution and pooling layer. Making \textbf{G}'s network deeper is theoretically a compelling way to capture more details in the image, however deep GANs are unstable and hard to train \cite{radford2015unsupervised, kodali2017convergence}. A trained generator learns the mapping \(G(z): z \longmapsto x\) from latent space representations \(z\) to realistic, \(2D\), chest X-ray images. 
\par
\begin{figure}
\centering
\includegraphics[width=80mm]{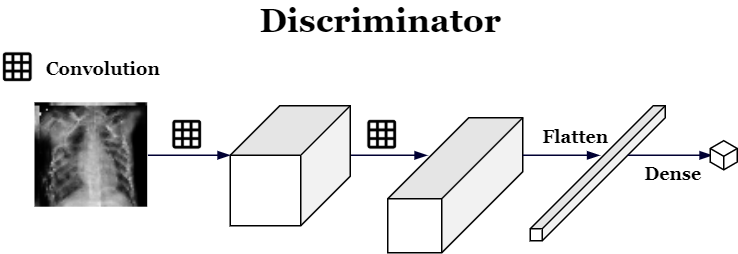}
\caption{Discriminator Architecture} \label{fig3}
\end{figure}

 The discriminator (\textbf{D}) (Figure~\ref{fig3}) is a 4-layer CNN that maps a 2D image to a scalar output that can be interpreted as the probability of the given input being a real chest X-ray image sampled from training data or image G(z) generated by the generator G. 
Optimization of D and G can be thought of as the following game of minimax \cite{goodfellow2016nips} with the value function \(V(G, D)\): 
\begin{equation} \label{eq:1}
\min_G \max_D V(D, G) = \mathbb{E}_{x_{{\sim_P}_{data{(x)}}}} [\log D(x)] + \mathbb{E}_{z_{{\sim_P}_{z{(z)}}}} [\log (1 - D(G(z)))]
\end{equation}
During training, generator G is trained to minimize the accuracy of discriminator D's ability in distinguishing between real and generated images while the discriminator is trying to maximize the probability of assigning real training images the "real" and generated images from G, "fake" labels. During the training, G improves at generating more realistic images while D gets better at correctly identifying between real and generated images.

\section{Datasets}

\subsection{Dataset I}

We used the publicly available chest X-ray dataset \cite{kermany2018labeled} with two categories of Normal (1,575 images) and Pneumonia (4,265 images). The images were in JPEG format and varied in size with pixel values in \([0 , 255]\) range. We resized all images to $128 \times 128$ pixels. Images were normalized to have [-1, 1] range for \(\tanh\) non-linearity activation in the IAGAN architecture. We use our bigger cohort (pneumonia) as the training class. 500 images from each class were randomly selected to evaluate the models' performance while the rest of the images were used for augmentation and training different models.
\subsection{Dataset II}
Covid-chestxray dataset~\cite{cohen2020covid} is an ongoing effort by Cohen~\emph{et al.} to make a public COVID-19 dataset of chest X-ray images with COVID-19 radiological readings. Wang~\emph{et al.} used covid-chestxray dataset, along with four other publicly available datasets and compiled the COVIDx~\cite{wang2020covid} dataset. With the number of images growing, many deep learning models are trained and tested on this public dataset~\cite{wang2020covid, ozturk2020automated, hemdan2020covidx}. At the time of this study, the COVIDx dataset is comprised of 8,066 normal, 5,559 pneumonia, and 589 COVID-19 images. The images are in RGB format with pixel range of \([0 , 255]\) and have various sizes. To train the generative models in this study, all images were converted to gray scale, resized to $128 \times 128$ pixels and normalized to have pixel intensities in the \([-1 , 1]\) range. 589 images from normal and pneumonia classes were randomly selected along with 589 COVID-19 images to test the models while the rest of the images were used for augmentation and training different models. 

\subsubsection{Segmentation of COVIDx Dataset}
\begin{figure}[h]
\centering
\includegraphics[width=67mm]{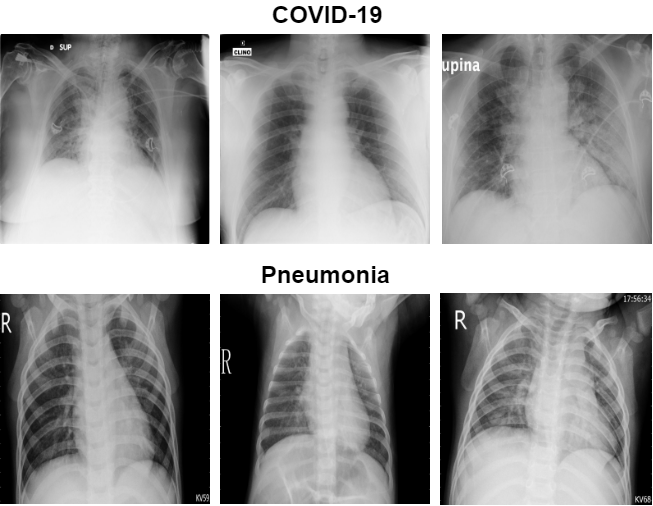}
\caption{Pneumonia and COVID-19 sample images from COVIDx dataset with class consistent annotations} \label{shortcut}
\end{figure}

A recent study \cite{degrave2020ai} using the COVIDx dataset showed that existing markers such as annotations and arrows outside of the lung on the X-ray images can act as shortcuts \cite{geirhos2020shortcut} in detecting COVID-19 using those shortcuts instead of actual COVID-19 disease markers. Figure \ref{shortcut} shows annotations on the top left of COVID-19 images which are consistent with the rest of the COVID-19 images and the \textit{R} symbol positioned on the left of pneumonia images consistent with images from the pneumonia class in COVIDx dataset.
\par To mitigate the effect of non-disease markers on our model, we segmented the lungs for the COVIDx dataset images. 900 randomly selected images (300 from each class) were manually segmented by an expert radiologist. A modified U-NET model \cite{motamed2019transfer}, pre-trained on the Montgomery chest X-ray dataset \cite{jaeger2014two} was fine-tuned using the 800 COVIDx segmentations. The segmentation model was tested on the 100 remaining ground truth images and achieved a Sørensen–Dice coefficient of 0.835.

\section{Data Augmentation}
\subsection{IAGAN}
We trained multiple instances of IAGAN outlined below. The architecture of IAGAN was kept unchanged for each instance and learning rates of 0.0004 and 0.0001 were used for the discriminator and generator, respectively. Experimenting with the size of the Gaussian noise vector z showed 120 to be the optimal size. We trained our IAGAN for 250 epochs on an Nvidia GeForce RTX 2080 Ti - 11 GB with a batch size of 32.
For dataset I, IAGAN was trained on 3,765 pneumonia images and tested on 500 pneumonia vs. 500 normal cases. For dataset II, one IAGAN was trained on 4,700 Pneumonia images and one IAGAN was trained on 7,477 Normal images. After successful training of the IAGAN, the generator has learned the distribution of the images of the training class. 
\par
\begin{figure}
 \centering
 \includegraphics[width= 12cm]{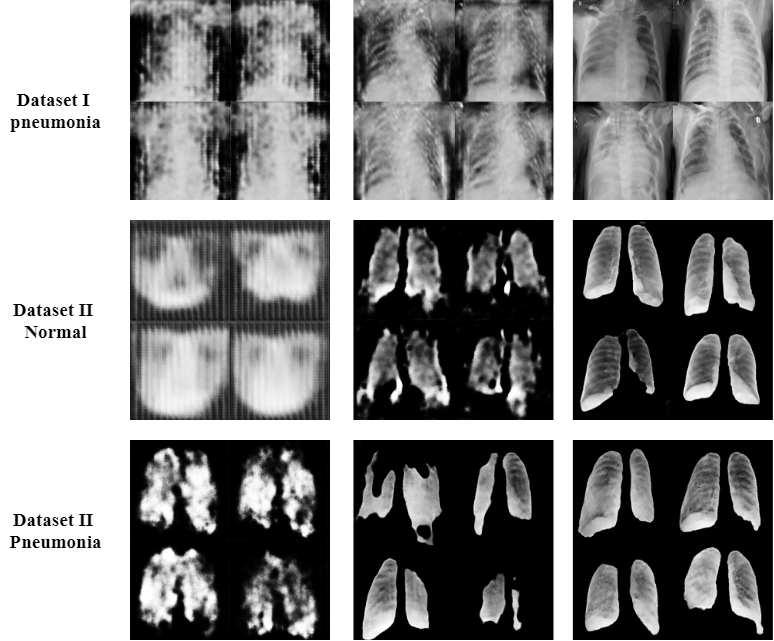}
 \caption{Generator's output during training}\label{iagan-epoch}
\end{figure}

 To generate new data, for each input image to IAGAN, 3 random noise vectors were initiated and 3 new images were generated from the generator. For dataset I, 3,765 pneumonia training images were put through G and for each image, three new images were generated (11,295). For each normal image that was not used for testing the model's performance, we did the same and generated 3,225 images from 1,075 normal images. Similarly, for dataset II, normal and pneumonia training images were put through the two trained generators, one generator from the IAGAN trained on normal images and one trained on pneumonia images. Similar to dataset I, each generator generated 3 new images using pneumonia and normal images that are not used in testing the model. Figure \ref{iagan-epoch} shows the generator's output at early, mid and later stages (from left to right respectively) of the training on datasets I and II.

Table \ref{table:split} shows the number of images for each class, before and after data augmentation using IAGAN. Dataset I does not have any COVID-19 images and does not use any normal images for training. Dataset II uses all COVID-19 images (589) for testing the model and hence, no augmentation is done using this class. Both normal and pneumonia class images are used for training the model and therefore, 589 randomly selected images are fixed to test the model from each class, the rest of the images are augmented using two separately trained IAGANs. One IAGAN trained on normal images, uses normal and pneumonia images to generated more normal images. The other IAGAN, uses normal and pneumonia images to generate more pneumonia images. 
\begin{table}[h!]
\begin{center}
 \begin{tabular}{||c c c c||} 
 \hline
 & Normal & Pneumonia & COVID-19 \\
 & (Training / Test) & (Training / Test) & (Training / Test)\\ [0.5ex] 
 \hline\hline

 Dataset I & 0/500 & 3,765/500 & N/A \\
 \hline
 Augmented Dataset I & 0/500 & 19,360/500 & N/A \\
 \hline
 Dataset II & 7,477/589 & 4,700/589 & 0/589 \\
 \hline
 Augmented Dataset II & 48,708/589 & 48,708/589 & 0/589 \\ [1ex] 
 \hline
 
\end{tabular}
\caption{IAGAN Augmentation}
 \label{table:split}
\end{center}
\end{table}

\subsection{DCGAN}
To understand the effect of our input image to IAGAN's generator, which allows using images from all classes to be fed into a trained generator for augmentation, we trained a DCGAN~\cite{radford2015unsupervised} that uses only the traditional Gaussian noise vector input to the generator. We used the same hyper-parameters and number of epochs as IAGAN. The only difference in the number of generated images is that images from classes other than what the DCGAN's Generator was trained on cannot be fed to the trained \textbf{G} for generating new images. For this reason, we generate 3 images for each image the DCGAN was trained on; for dataset I, 3 images were generated for each pneumonia training image (3 similar images were generated using the anomaly score defined by Schlegl et. al\cite{schlegl2017unsupervised} and for dataset II, two DCGANs were trained similar to IAGAN, 3 images were generated for each normal training image with the \textbf{G} trained on normal images and 3 images were generated for each pneumonia training images with the \textbf{G} trained on pneumonia images. Table \ref{table:dcaug} shows the number of images for each class, before and after data augmentation using DCGAN .  
\begin{table}[h!]
\begin{center}
 \begin{tabular}{||c c c c||} 
 \hline
 & Normal & Pneumonia & COVID-19 \\
 & (Train / Test) & (Train / Test) & (Train / Test)\\ [0.5ex] 
 \hline\hline
 Augmented Dataset I & 0/500 & 15,060/500 & N/A \\
 \hline
 Augmented Dataset II & 29,908/589 & 18,800/589 & 0/589 \\ [1ex] 
 \hline
\end{tabular}
\caption{DCGAN Augmentation}
 \label{table:dcaug}
\end{center}
\end{table}
\subsection{Traditional Augmentation}
\begin{figure}[h!]
 \centering
 \includegraphics[width= 11cm]{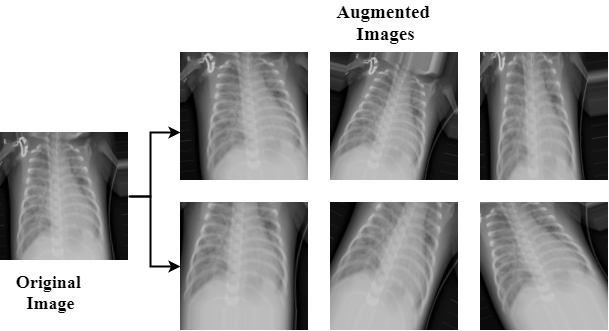}
 \caption{Traditional augmentation output sample}\label{trad-sample}
\end{figure}

Based on recent literature on data augmentation for chest X-ray pathology classification using CNNs \cite{stirenko2018chest}, we used Keras' data generator function for data augmentation by using random rotations in the range of 20 degrees, width and height shift in the range of 0.2 and zoom in the range of 0.2. For each training image, 8 new images were randomly generated using the aforementioned augmentation methods. Figure \ref{trad-sample} shows the sample output of this function. Table \ref{table:tradaug} shows the number of images for each class, before and after data augmentation using traditional augmentation methods.

\begin{table}[h!]
\begin{center}
 \begin{tabular}{||c c c c||} 
 \hline
 & Normal & Pneumonia & COVID-19 \\
 & (Train / Test) & (Train / Test) & (Train / Test)\\ [0.5ex] 
 \hline\hline
 Augmented Dataset I & 0/500 & 33,885/500 & N/A \\
 \hline
 Augmented Dataset II & 67,293/589 & 42,300/589 & 0/589 \\ [1ex] 
 \hline
\end{tabular}
\caption{Traditional Augmentation}
 \label{table:tradaug}
\end{center}
\end{table}

\section{Experiments}
Schlegl \emph{et al.} \cite{schlegl2017unsupervised} proposed AnoGAN for detecting anomalies in optical coherence tomography images of the retina. The AnoGAN architecture follows DCGAN~\cite{radford2015unsupervised} in terms of overall generator and discriminator architecture. They trained the AnoGAN model on one class of images. With the trained generator \textbf{G} at hand, in order to find anomalies in test image x, back-propagation (using equation \ref{eq:4} with \(\lambda = 0.2\)) was used to find a point \(z_i\) that generates an image that looks similar to x. Upon finding a point z after a set number of iterations (800 iterations in our experiments), the anomaly score \(A(x)\) (equation \ref{eq:5}) is defined using residual and discrimination losses as shown below, calculated at point \(z\). \(L_R\) and \(L_D\) are the residual and discriminator loss that enforce visual and image characteristic similarity between real image x and generated image \(G(z_i)\). The discriminator loss captures image characteristics using the output of an intermediate layer of the discriminator, $f(.)$, making the discriminator act as an image encoder.
\begin{equation}\label{eq:2}
\mathcal{L}_R({z_i}) = \sum|x - G(z_i)|
\end{equation}
\begin{equation}\label{eq:3}
\mathcal{L}_D({z_i}) = \sum|f(x) - f(G(z_i))|
\end{equation}
\begin{equation}\label{eq:4}
  \mathcal{L}({z_i}) = (1 - \lambda) \times \mathcal{L}_{R}({z_i}) + \lambda \times \mathcal{L}_{D}({z_i})
\end{equation}
\begin{equation} \label{eq:5}
A(x) = (1 - \lambda) \times \mathcal{L}_{R}({z}) + \lambda \times \mathcal{L}_{D}({z})
\end{equation}
\subsection{Dataset I}
We used the AnoGAN architecture to evaluate the effects of different approaches to data augmentation. We trained 4 AnoGAN models; one trained on pneumonia images from dataset I and the other 3 were trained on augmented pneumonia images with IAGAN, DCGAN and traditional augmentation methods. 
\subsection{Dataset II}
To detect COVID-19 positive from COVID-19 negative images, one AnoGAN was trained on normal images and another identical network was trained on pneumonia images. After calculating two anomaly scores for each test image, one calculated by each AnoGAN, the sum of two anomaly scores was assigned as the final anomaly score for the test image. The idea is that the AnoGAN trained on normal images will result in lower anomaly score for normal images during test while AnoGAN trained on pneumonia images results in lower scores for pneumonia images. In both networks, the COVID-19 images produce higher anomaly scores hence the COVID-19 final anomaly score will be higher than the normal and pneumonia classes.
\par The AnoGAN pair model were trained similar to AnoGAN on dataset I; trained on normal and pneumonia training images without augmentation, normal and pneumonia images augmented using IAGAN, DCGAN and traditional augmentation methods. 
\section{Results}
We calculated the area under the ROC curve (AUC) for each model trained on datasets I and II, before and after data augmentation. For dataset I, AUC represents the classification capability of detecting pneumonia vs. normal chest X-rays. For dataset II, we classify COVID-19 positive from COVID-19 negative images. With 589 test images from each class (normal, pneumonia and COVID-19) in dataset II, we calculated the AUC for the balanced COVID-19 negative class vs. COVID-19 positive test images. The balanced COVID-19 negative class was created by randomly sampling 294 normal and 295 pneumonia images from 589 normal and 589 pneumonia test images. 

Table \ref{table:auc} shows the calculated AUC for datasets I and II. It can be seen that our proposed IAGAN augmentation method outperforms all other three models for both Dataset I and II: no augmentation, DCGAN, and traditional augmentation methods. DeLong test \cite{delong1988comparing} was used to compare the AUC of the models by calculating the \textit{p-value} for significance difference. The \textit{p-values} are added next to the AUC of each augmentation method and measures the significance of the model compared to the model trained with no augmentation.

\begin{table}[h!]
\begin{center}
 \begin{tabular}{||c c c c c||} 
 \hline
 & No Augmentation & IAGAN & DCGAN & Traditional Augmentation \\ [0.5ex] 
 \hline\hline
 
 Dataset I & 0.87 & \textbf{0.90} (p = \(3.17 \times 10 ^ {-7}\)) & 0.87 (p = 0.5) & 0.88 (p = 0.08) \\
 \hline
 Dataset II & 0.74 & \textbf{0.76} (p = 0.01) & 0.75 (p = 0.43) & 0.75 (p = 0.57)\\
 \hline
\end{tabular}
\caption{AUC and p-value for datasets I and II }
 \label{table:auc}
\end{center}
\end{table}

\par We calculated the accuracy of each model at the highest sensitivity / specificity pair points (with minimum 0.80 sensitivity and specificity) for each model trained on datasets I and II. Table \ref{table:acc} shows the sensitivity, specificity and accuracy of different trained models on both datasets where it can be seen that our proposed IAGAN outperforms all other models in both sensitivity and specificity.

\begin{table}[h!]
\begin{center}
 \begin{tabular}{||c c c c||} 
 \hline
 Model (Datasets I / II) & Sensitivity & Specificity & Accuracy \\ [0.5ex] 
 \hline\hline
 No augmentation  & 0.80 / 0.67 & 0.81 / 0.68 & 0.80 / 0.67 \\
 \hline
 IAGAN & \textbf{0.82 / 0.69}   &\textbf{0.84 / 0.69} & 0.80 / \textbf{0.69} \\
 \hline
 DCGAN  & 0.80 / 0.67 & 0.81 / 0.67 & 0.80 / 0.67 \\
 \hline
 Traditional augmentation & 0.80 / 0.68 & 0.81 / 0.68 & 0.80 / 0.68 \\
 \hline
\end{tabular}
\\
\caption{Sensitivity, Specificity and Accuracy for datasets I and II, respectively}
 \label{table:acc}
\end{center}
\end{table}

\section{Discussion}
Harnessing GANs' ability to generate never-before-seen data, by learning the distribution of images, allows for augmentation of data that is not limited to applying different transformations to existing images. By using the proposed IAGAN, not only are we able to generate new images for the same class used to augment data (e.g., using normal images to augment normal dataset), but also generate new images of any class withing that domain of images using one class of images (e.g., generating chest X-rays with pneumonia, COVID-19 or healthy cases using normal images).
\par We showed that a traditional DCGAN with a single random noise vector input to the generator fails to effectively augment data for a GAN. Traditional augmentation methods showed improved prediction in a subset of the tasks (AUC of 0.75 vs 0.74 for dataset II), yet failed to effectively improve the accuracy of the overall models with statistical significance. Our proposed IAGAN architecture, however, improves the models' accuracy when used for augmentation of the training cohort, with statistical significance. We used the AnoGAN \cite{schlegl2017unsupervised} architecture to show when the training data is augmented using our proposed IAGAN method, the AUC improves by \(3\%\) and \(2\%\), compared to no augmentation, for dataset I and II, respectively. IAGAN also showed improved sensitivity / specificity for the AnoGAN model (\(2\% - 3\%\) for dataset I and \(2\% - 1\%\) for dataset II in sensitivity and specificity respectively).  
\par IAGAN architecture allows for semi-supervised augmentation of data for a specific class of labels. We showed that by training IAGAN on a specific class, we were able to use all classes to generate new data for that specific class. Effective training of generative models for medical imaging can be specially helpful to detect anomalies in classes where we do not have enough data / labels for effectively training CNN models. The COVID-19 pandemic is a great example for the importance of generative models, where no images are required for this class of images in order to detect images of this class~\cite{Motamed2020}. Advances in generative models for detection of anomalies can allow for fast deployment of such models at a time where adequate number of labelled images for the new disease are not available for the effective training of CNNs. It is worth mentioning that while an architecture like CycleGAN \cite{zhu2017unpaired} uses images as input to its generator, to train a CycleGAN, images from two different domain (i.e normal and pneumonia) are used to learn the transition of one image domain to the other. While this could allow for augmenting data from one class to the other, it would require having enough labelled data for all classes and does not allow for single class data augmentation (i.e augmenting normal dataset using partially labelled chest X-rays with only available label being normal) as is enabled by IAGAN.

\begin{figure}[h]
\centering
\includegraphics[width=120mm]{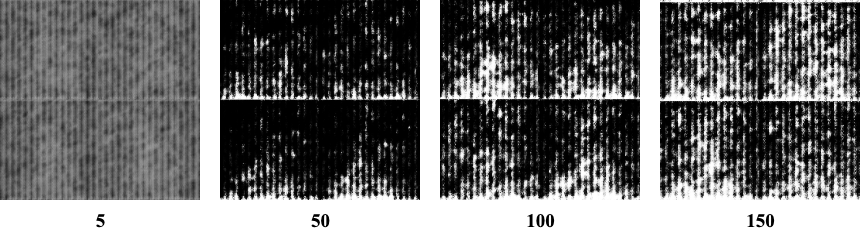}
\caption{IAGAN's generator output at different epochs of the model training with random generated input images} \label{rand-input}
\end{figure}

Early on in this study, it was not immediately clear whether the effects of feeding real images to GAN's generator (\textbf{G}) was due to image specific information, or providing the model with a larger vector size in the generator's up-sampling path. Since the down-sampled image is concatenated with \textbf{G}'s other input early on in the network, the effects of the input image might be associated with the added vector size, having the same effect as adding the same image with randomly sampled pixel valued. We trained the IAGAN but this time, the input images were randomly generated. The IAGAN failed to generate realistic images using random input images. This confirms that our proposed IAGAN architecture that encodes the input images using convolution and attention layers to a lower-dimensional representation, before concatenating with the projected noise is an effective way to generate meaningful images and augment data. Figure \ref{rand-input} shows \textbf{G}'s output in epochs \(5-150\). 

\par One of the disadvantages of using a dataset such as COVIDx, compared to dataset I, is the multicentric nature of the images. Since images have been collected from multiple sources and health centers with possibly different acquisition parameters and different scanner models, we observed that our GAN for anomaly detection does not perform as well as dataset I, with or without augmentation. With a more consistent dataset, we hope to achieve improved results on dataset II, compared to dataset I.
\section{Conclusion}
In this paper, we presented IAGAN; a semi-supervised GAN-based augmentation method to improve training GANs for detection of anomalies (pneumonia and COVID-19) in chest X-rays. IAGAN showed to be statistically significant in augmenting data, improving the AUC, sensitivity and specificity of GAN for detection of anomalies.
\section{Acknowledgements}
This research was funded by Chair in Medical Imaging and Artificial Intelligence, a joint Hospital-University Chair between the University of Toronto, The Hospital for Sick Children, and the SickKids Foundation. 
\bibliography{references}

\begin{thebibliography}{10}

\bibitem{krizhevsky2012imagenet}
Alex Krizhevsky, Ilya Sutskever, and Geoffrey~E Hinton.
\newblock Imagenet classification with deep convolutional neural networks.
\newblock In {\em Advances in neural information processing systems}, pages
  1097--1105, 2012.

\bibitem{he2015delving}
Kaiming He, Xiangyu Zhang, Shaoqing Ren, and Jian Sun.
\newblock Delving deep into rectifiers: Surpassing human-level performance on
  imagenet classification.
\newblock In {\em Proceedings of the IEEE international conference on computer
  vision}, pages 1026--1034, 2015.

\bibitem{he2016deep}
Kaiming He, Xiangyu Zhang, Shaoqing Ren, and Jian Sun.
\newblock Deep residual learning for image recognition.
\newblock In {\em Proceedings of the IEEE conference on computer vision and
  pattern recognition}, pages 770--778, 2016.

\bibitem{zhang2019image}
Yu-Dong Zhang, Zhengchao Dong, Xianqing Chen, Wenjuan Jia, Sidan Du, Khan
  Muhammad, and Shui-Hua Wang.
\newblock Image based fruit category classification by 13-layer deep
  convolutional neural network and data augmentation.
\newblock {\em Multimedia Tools and Applications}, 78(3):3613--3632, 2019.

\bibitem{hao2020comprehensive}
Ruqian Hao, Khashayar Namdar, Lin Liu, Masoom~A. Haider, and Farzad Khalvati.
\newblock A comprehensive study of data augmentation strategies for prostate
  cancer detection in diffusion-weighted mri using convolutional neural
  networks.
\newblock {\em arXiv preprint arXiv.2006.01693}, 2020.

\bibitem{goodfellow2016nips}
Ian Goodfellow.
\newblock Nips 2016 tutorial: Generative adversarial networks.
\newblock {\em arXiv preprint arXiv:1701.00160}, 2016.

\bibitem{sandfort2019data}
Veit Sandfort, Ke~Yan, Perry~J Pickhardt, and Ronald~M Summers.
\newblock Data augmentation using generative adversarial networks (cyclegan) to
  improve generalizability in ct segmentation tasks.
\newblock {\em Scientific reports}, 9(1):1--9, 2019.

\bibitem{ronneberger2015u}
Olaf Ronneberger, Philipp Fischer, and Thomas Brox.
\newblock U-net: Convolutional networks for biomedical image segmentation.
\newblock In {\em International Conference on Medical image computing and
  computer-assisted intervention}, pages 234--241. Springer, 2015.

\bibitem{radford2015unsupervised}
Alec Radford, Luke Metz, and Soumith Chintala.
\newblock Unsupervised representation learning with deep convolutional
  generative adversarial networks.
\newblock {\em arXiv preprint arXiv:1511.06434}, 2015.

\bibitem{mirza2014conditional}
Mehdi Mirza and Simon Osindero.
\newblock Conditional generative adversarial nets.
\newblock {\em arXiv preprint arXiv:1411.1784}, 2014.

\bibitem{frid2018gan}
Maayan Frid-Adar, Idit Diamant, Eyal Klang, Michal Amitai, Jacob Goldberger,
  and Hayit Greenspan.
\newblock Gan-based synthetic medical image augmentation for increased cnn
  performance in liver lesion classification.
\newblock {\em Neurocomputing}, 321:321--331, 2018.

\bibitem{wu2018conditional}
Eric Wu, Kevin Wu, David Cox, and William Lotter.
\newblock Conditional infilling gans for data augmentation in mammogram
  classification.
\newblock In {\em Image Analysis for Moving Organ, Breast, and Thoracic
  Images}, pages 98--106. Springer, 2018.

\bibitem{antoniou2017data}
Antreas Antoniou, Amos Storkey, and Harrison Edwards.
\newblock Data augmentation generative adversarial networks.
\newblock {\em arXiv preprint arXiv:1711.04340}, 2017.

\bibitem{schlegl2017unsupervised}
Thomas Schlegl, Philipp Seeb{\"o}ck, Sebastian~M Waldstein, Ursula
  Schmidt-Erfurth, and Georg Langs.
\newblock Unsupervised anomaly detection with generative adversarial networks
  to guide marker discovery.
\newblock In {\em International conference on information processing in medical
  imaging}, pages 146--157. Springer, 2017.

\bibitem{zhang2018self}
Han Zhang, Ian Goodfellow, Dimitris Metaxas, and Augustus Odena.
\newblock Self-attention generative adversarial networks.
\newblock {\em arXiv preprint arXiv:1805.08318}, 2018.

\bibitem{szegedy2016rethinking}
Christian Szegedy, Vincent Vanhoucke, Sergey Ioffe, Jon Shlens, and Zbigniew
  Wojna.
\newblock Rethinking the inception architecture for computer vision.
\newblock In {\em Proceedings of the IEEE conference on computer vision and
  pattern recognition}, pages 2818--2826, 2016.

\bibitem{kodali2017convergence}
Naveen Kodali, Jacob Abernethy, James Hays, and Zsolt Kira.
\newblock On convergence and stability of gans.
\newblock {\em arXiv preprint arXiv:1705.07215}, 2017.

\bibitem{kermany2018labeled}
Daniel Kermany, Kang Zhang, and Michael Goldbaum.
\newblock Labeled optical coherence tomography (oct) and chest x-ray images for
  classification.
\newblock {\em Mendeley data}, 2, 2018.

\bibitem{cohen2020covid}
Joseph~Paul Cohen, Paul Morrison, and Lan Dao.
\newblock Covid-19 image data collection. arxiv 2003.11597, 2020.
\newblock {\em URL https://github. com/ieee8023/covid-chestxray-dataset}, 2020.

\bibitem{wang2020covid}
Linda Wang and Alexander Wong.
\newblock Covid-net: A tailored deep convolutional neural network design for
  detection of covid-19 cases from chest x-ray images.
\newblock {\em arXiv preprint arXiv:2003.09871}, 2020.

\bibitem{ozturk2020automated}
Tulin Ozturk, Muhammed Talo, Eylul~Azra Yildirim, Ulas~Baran Baloglu, Ozal
  Yildirim, and U~Rajendra Acharya.
\newblock Automated detection of covid-19 cases using deep neural networks with
  x-ray images.
\newblock {\em Computers in Biology and Medicine}, page 103792, 2020.

\bibitem{hemdan2020covidx}
Ezz El-Din Hemdan, Marwa~A Shouman, and Mohamed~Esmail Karar.
\newblock Covidx-net: A framework of deep learning classifiers to diagnose
  covid-19 in x-ray images.
\newblock {\em arXiv preprint arXiv:2003.11055}, 2020.

\bibitem{degrave2020ai}
Alex~J DeGrave, Joseph~D Janizek, and Su-In Lee.
\newblock Ai for radiographic covid-19 detection selects shortcuts over signal.
\newblock {\em medRxiv}, 2020.

\bibitem{geirhos2020shortcut}
Robert Geirhos, J{\"o}rn-Henrik Jacobsen, Claudio Michaelis, Richard Zemel,
  Wieland Brendel, Matthias Bethge, and Felix~A Wichmann.
\newblock Shortcut learning in deep neural networks.
\newblock {\em arXiv preprint arXiv:2004.07780}, 2020.

\bibitem{motamed2019transfer}
Saman Motamed, Isha Gujrathi, Dominik Deniffel, Anton Oentoro, Masoom~A Haider,
  and Farzad Khalvati.
\newblock A transfer learning approach for automated segmentation of prostate
  whole gland and transition zone in diffusion weighted mri.
\newblock {\em arXiv preprint arXiv:1909.09541}, 2019.

\bibitem{jaeger2014two}
Stefan Jaeger, Sema Candemir, Sameer Antani, Y{\`\i}-Xi{\'a}ng~J W{\'a}ng,
  Pu-Xuan Lu, and George Thoma.
\newblock Two public chest x-ray datasets for computer-aided screening of
  pulmonary diseases.
\newblock {\em Quantitative imaging in medicine and surgery}, 4(6):475, 2014.

\bibitem{stirenko2018chest}
Sergii Stirenko, Yuriy Kochura, Oleg Alienin, Oleksandr Rokovyi, Yuri
  Gordienko, Peng Gang, and Wei Zeng.
\newblock Chest x-ray analysis of tuberculosis by deep learning with
  segmentation and augmentation.
\newblock In {\em 2018 IEEE 38th International Conference on Electronics and
  Nanotechnology (ELNANO)}, pages 422--428. IEEE, 2018.

\bibitem{delong1988comparing}
Elizabeth~R DeLong, David~M DeLong, and Daniel~L Clarke-Pearson.
\newblock Comparing the areas under two or more correlated receiver operating
  characteristic curves: a nonparametric approach.
\newblock {\em Biometrics}, pages 837--845, 1988.

\bibitem{Motamed2020}
Saman Motamed, Patrik Rogalla, and Farzad Khalvati.
\newblock Randgan: Randomized generative adversarial network for detection of
  covid-19 in chest x-ray.
\newblock {\em arXiv}, 2020.

\end{thebibliography}
\bibliographystyle{unsrt}
\end{document}